\documentclass{article}

\usepackage{arxiv}
\usepackage{graphicx}
\usepackage{latexsym}

\usepackage{times}
\usepackage{url}
\usepackage[hidelinks]{hyperref}
\usepackage[utf8]{inputenc}
\usepackage[small]{caption}
\usepackage{graphicx}
\usepackage{subfig}
\usepackage{amsmath}
\usepackage{amsthm}
\usepackage{amssymb}
\usepackage{booktabs}
\usepackage{algorithm}
\usepackage{algorithmic}
\usepackage[switch]{lineno}

\usepackage[square,numbers]{natbib}

\hypersetup{
    colorlinks=true,
    linkcolor=blue,
    filecolor=magenta,      
    urlcolor=cyan,
    pdfborderstyle={/S/U}
}

\title{A statistical approach to detect sensitive features in a group fairness setting}

\author{ \href{https://orcid.org/0000-0001-7301-6167}{\includegraphics[scale=0.06]{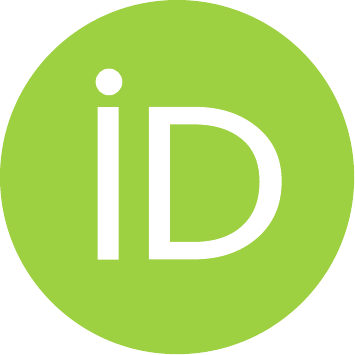}\hspace{1mm}Guilherme Dean Pelegrina} \\
	School of Applied Sciences (FCA)\\
	University of Campinas (UNICAMP)\\
	Limeira, Brazil \\
	\texttt{guidean@unicamp.br} \\
	\And
	\href{https://orcid.org/0000-0003-2316-7623}{\includegraphics[scale=0.06]{orcid.pdf}\hspace{1mm}Miguel Couceiro} \\
	Université de Lorraine, CNRS, LORIA\\
	F-54000 Nancy, France \\
	\texttt{miguel.couceiro@loria.fr} \\
 \And
	\href{https://orcid.org/0000-0003-0290-0080}{\includegraphics[scale=0.06]{orcid.pdf}\hspace{1mm}Leonardo Tomazeli Duarte} \\
	School of Applied Sciences (FCA)\\
	University of Campinas (UNICAMP)\\
	Limeira, Brazil \\
	\texttt{leonardo.duarte@fca.unicamp.br} \\
}

\begin{document}
\maketitle

\begin{abstract}
    The use of machine learning models in decision support systems with high societal impact raised concerns about unfair (disparate) results for different groups of people. 
    When evaluating such unfair decisions, one generally relies on predefined groups that are determined by a set of features that are considered sensitive. However, such an approach is subjective and does not guarantee that these features are the only ones to be considered as sensitive nor that they entail unfair (disparate) outcomes. 
    
    In this paper, we propose a preprocessing step to address the task of automatically recognizing sensitive features that does not require a trained model to verify unfair results. Our proposal is based on the Hilbert–Schmidt independence criterion, which measures the statistical dependence of variable distributions. We hypothesize that if the dependence between the label vector and a candidate is high for a sensitive feature, then the information provided by this feature will entail disparate performance measures between  groups. Our empirical results attest our hypothesis and show that several features considered as sensitive in the literature do not necessarily entail disparate (unfair) results.  
\end{abstract}

\section{Introduction}

The use of Machine Learning (ML) methods to deal with real-world problems has grown exponentially in the last decades~\cite{Brink2016,Sarker2021}. 
In particular, ML is currently integrated in every decision support (DS) system with critical societal impacts. This fact has raised increasing concerns about the (un)fairness of such `artificial deciders' in several contexts~\cite{Mehrabi2019}, such as in recidivism risk prediction~\cite{Angwin2016}, facial recognition systems~\cite{Raji2019}, platforms for job applications~\cite{hangartner2021monitoring}, hate speech and abusive language detection \cite{davidson2019racial}. This motivated several efforts towards both the detection  and mitigation of unfairness in ML and DSs. The former are mostly centered in the detection of undesirable bias, while the later seek to reduce such biases. These methods fall into four main categories, namely, pre-processing \cite{calders2009building,roh2020fairbatch,Pelegrina2022,Pelegrina2022b}, in-processing \cite{zhang2018mitigating,agarwal2018reductions,iosifidis2019adafair}, post-processing \cite{hardt2016equality,fish2016confidence,BhargavaCN20}, and  hybrid-processing\footnote{These methods combine different fairness interventions such as in-processing and post-processing.} \cite{zemel2013learning,iosifidis2019fae,alves2021reducing}.

Fairness of decision systems is commonly assessed via metrics  that rely on the outcomes\footnote{Some works also focus on procedural fairness, {\it i.e.}, of the decision making processes (means) that lead to the outcomes \cite{GummadiProcess18}. However, this concern will remain outside the scope of this paper.} of decision models, and  inspired by anti-discrimination laws {\it under which decision policies or practices can be declared as discriminatory based on their effects on people belonging to certain sensitive demographic groups} \cite{Gummadi2016}, {\it e.g.}, gender, race, age and sexual orientation.  Still, fairness is a social construct~\cite{jacobs2021measurement} and an ethical concept~\cite{tsamados2022ethics}, and its definition  remains prone to subjectivity.

To overcome subjectivity, several fairness notions have been recently proposed to objectively measure and to capture different aspects
of fairness. These include group based notions ({\it e.g.}, \cite{dwork2012fairness,hardt2016equality}), individual-based notions ({\it e.g.}, \cite{dwork2012fairness,Gummadi18}), and causal and counterfactual based notions  ({\it e.g.},  \cite{pearl2009causality,kusner2017counterfactual}). Even though these metrics would replace human subjectivity by objective metrics, they are still dependent on subjective decisions. 

Indeed, most of the above metrics rely on the choice of particular features (variables) with respect to which fairness will be measured. For instance, in the case of group based metrics, fairness is mostly measured with respect to disparate performance and decision results among different subpopulations. For example, when predicting recidivism risk~\cite{Angwin2016}, disparate results among the white and black subpopulations, or among men and women.    
In most of the fairness analysis in ML and DS problems, it is assumed that the population is divided into privileged and unprivileged groups on which disparate results are expected. In other words, the {\it sensitive features} that discriminate these groups are known {\it a priori}. Examples of commonly considered sensitive features include race, gender, age, sexual orientation, etc. See, {\it e.g.},~\cite{LeQuy2022} for a set of ML problems and the sensitive features frequently considered in the literature.

However, the choice of sensitive features also remains subjective and, as they are chosen {\it a priori}, it is dataset agnostic and does not necessarily entail outcome disparities in a given use case scenario. Indeed, consider an ML problem in which both gender and race are among the set of features. Depending on problem addressed, one may have uneven results with respect to race but similar results regardless the gender. In this situation, one may only be aware of race as a sensitive feature when training the ML model. As further illustration, consider one of the above mentioned platforms for job applications. Ethically, physical disabilities of candidates should not constitute a discriminating ({\it i.e.}, it should be considered sensitive) of candidates. However, if the job is indeed of physical nature, then some physical abilities may be required and, therefore, this information should not be considered as sensitive.

In this paper, we address this critical issue of identifying sensitive feature or subfeatures ({\it i.e.}, corresponding protected groups or subgroups), and we propose to defining them as {\it features that entail disparate (unfair) outcomes}.  
However, in order to verify disparities among different (sub)groups, one usually needs to train the ML model and calculate the pertaining fairness metrics. This step may be computationally costly depending on the ML model is adopted.
Thus this raises the question: how to detect possible disparities before the training step? 

To tackle this issue, we propose a statistical approach based on the Hilbert–Schmidt independence criterion (HSIC)~\cite{Gretton2005}, which is commonly used in data science ~\cite{Wang2021} to measure the dependence between two matrices (or vectors). The HSIC has been used as a feature selection method~\cite{Song2007,Song2012}, for instance, in classification or regression tasks. The HSIC is usually used under the assumption that ``good features'' are those that maximize the dependence degree between these features and the vector of labels or values, and  one selects the subset of features with the highest dependence degrees, as they are more relevant to the output. In~\cite{Barshan2011}, the authors used the empirical estimate of HSIC in order to derive a supervised version of Principal Component Analysis (PCA). The idea was to perform dimensionality reduction while maximizing the dependence between the projected data and the vector of outcomes. HSIC has also been used when defining the loss-function for regression of classification task~\cite{Wang2020,Greenfeld2020}. In this context, a noteworthy application addresses fairness in automatic decisions by learning models that reduce disparities between sensitive groups~\cite{Perez-Suay2017,Li2022}.

Our hypothesis in this work is that, for a given feature, if 
\begin{enumerate}
\item this feature brings information that splits people into different groups, and
\item the HSIC\footnote{In fact, as will be further discussed in this paper, we adopted a normalized version of HSIC.} between this feature and vector of labels is high,
\end{enumerate}
then {\it there will be disparate outcomes among the  (sub)groups discriminated by this feature}. The intuition is that the features with high HSIC are important to the classification task as they carry key information to discriminate (sub)classes. Hence, it is likely to observe disparate outcomes among the discriminated (sub)groups. 

Note that not all features that split the population into two groups can be considered as sensitive. For example, the charge degree in recidivism risk prediction~\cite{Angwin2016} may split the population between those who committed a felony or a misdemeanor, but it is not assumed as a sensitive one. Moreover, sensitive features may split the population into more than two groups. For instance, as race, one may have whites, blacks and Asians. For each subfeature ({\it e.g.}, race\_whites, race\_blacks and race\_Asians), a high HSIC will indicate (i) if the feature is sensitive and (ii) which group division may lead to unfair (disparate) results. 
 
As we will see, HSIC indicates both the sensitive features and the associate groups of individuals that may lead to unfair outcomes. More precisely,  high HSIC values  between features (or subfeatures, in the case of categorical data) and the label vector may entail disparate results. We attest our hypothesis with empirical evaluations on several datasets frequently used in the literature, and that relate high HSIC values to high group fairness measures. 
This HSIC based preprocessing approach thus constitutes a noteworthy tool  researchers and practitioners to automatically detect those features that should be considered sensitive in an ML or DSs tasks.

It is important to stress that our approach deviates from previous attempts such as \cite{Gummadi2016,GummadiProcess18} in two major aspects, namely, by seeking an automatic approach to choosing which features should be considered sensitive, and by claiming that the choice of sensitive features should take into account both the task and use case scenario. In particular, we show that   certain features often considered as sensitive from ethical and social standpoints do not result in outcome disparaties, and thus their use in ML and DSs should not be prohibited. 
In fact, our approach also enables proxy detection as we will discuss in Section~\ref{sec:exper}.

The rest of this paper is organized as follows. In Section~\ref{sec:notation}, we outline the notations used in this paper. Section~\ref{sec:hsic} discusses the Hilbert–Schmidt independence criterion and its normalized version. The proposed approach to automatically detect sensitive features and sensitive groups is presented in Section~\ref{sec:proposal}. In Section~\ref{sec:exper}, we conduct the numerical experiments and discuss the obtained results. Finally, in Section~\ref{sec:concl}, we present our conclusions and future perspectives. 

\paragraph{Main contributions.}
Here, we simply highlight the main contributions of the paper.
\begin{itemize}
\item We address the problem of detecting sensitive features in group fairness settings, and propose a statistical approach based on the HSIC that does not require a trained model to verify disparate outcomes.
\item We present preliminary empirical results on four well known datasets that support our hypothesis. The higher is the HSIC-based dependence measure between a feature and the labels, the higher are the disparate results entailed by using the information provided by such a feature.
\item The analysis of our results also shows that several features taken as sensitive in the literature do not necessarily entail disparate (unfair) results. 
\item We also raise some perspectives for future work. Our proposed approach can be easily extended to decide whether a combination features ({\it e.g.}, the use of features describing gender and race) can lead to disparate results. Moreover, we will investigate how to adapt our framework to multiclass classification problems and to individual fairness settings. 
\end{itemize}




\section{Notation}
\label{sec:notation}

Throughout this paper, we assume an underlying binary classification problem such that each $m$-dimensional sample is assigned to a class $y \in \left\{-1,1 \right\}$. The set of $n$ samples and the associated vector of classes are represented by $\mathbf{X}$ and $\mathbf{y}$, respectively. Each column of $\mathbf{X}$, defined by $\mathbf{X}^{(j)}$, describes a feature $ G_{j}$, $j=1, \ldots, m$.

Very often some features in ML problems are categorical. However, as most of the learning algorithms require numbers as input features, we consider in this paper the one-hot encoding method to transform the categorical features into binary ones. Mathematically, a vector $\mathbf{X}^{(j)}$ describing a categorical feature $G_{j}$ with feature space $\left\{G_{j,1}, G_{j,2}, \ldots, G_{j,q} \right\}$ can be replaced by\footnote{One generally replaces by $q-1$ binary features in order to avoid a redundant column. However, in our analysis, the use of either $q$ or $q-1$ binary features lead to the same results.} $q$ binary vectors $\mathbf{X}^{(j,k)}$, $k=1, \ldots, q$, that indicate which category each sample belongs to. We illustrate this scenario with the 3-dimensional dataset
\begin{equation}
\label{eq:lat_data}
\mathbf{X}=\left[\begin{array}{ccc}
male & 2 & \text{30-60} \\
female & 0 & <30 \\ 
female & 1 & >60
\end{array}
\right],
\end{equation}
whose features are 
\begin{itemize}
    \item[]$G_1$: $gender$ with possible values $male$ or $female$,
    \item[]$G_2$: number $ncrimes$ of committed crimes in the last 5 years, and 
    \item[]$G_3$: $age$ that can be less than 30, between 30 and 60 or greater than 60 years old.
\end{itemize}
    After encoding the categorical features, one achieves the extended 5-dimensional dataset
\begin{equation}
\label{eq:lat_data}
\tilde{\mathbf{X}}=\left[\begin{array}{cccccc}
1 & 0 & 2 & 0 & 1 & 0 \\
0 & 1 & 0 & 1 & 0 & 0 \\
0 & 1 & 1 & 0 & 0 & 1 \\
\end{array}
\right],
\end{equation}
where $G_{1,1} = gender_{male}$, $G_{1,2} = gender_{female}$, $G_{3,1} = age_{\leq30}$, $G_{3,2} = age_{30-60}$ and $G_{3,3} = age_{\geq60}$ are the novel subfeatures. Note that, after conducting one-hot encoding for all categorical features, one increases the dataset dimension from $m$ to $\tilde{m}$.

\section{Hilbert–Schmidt independence criterion}
\label{sec:hsic}

There are several measures adopted in feature selection tasks to evaluate the relation between inputs and outputs~\cite{Kotsiantis2011}. As examples, one may cite the correlation~\cite{Hall1999} and mutual information~\cite{Vergara2014} between variables. Methods based on correlation are generally straightforward to be deployed as this measure is easy to be calculated. However, correlation only provides second-order information between variables and, therefore, does not necessarily imply dependence. On the other hand, mutual information brings the knowledge about the dependence degree between variables. The price to be paid here is that, in order to calculate the mutual information, one needs to estimate the probability density function of such variables. This task is very impractical in most cases.

Another measure that has been used in recent works in the literature is the Hilbert–Schmidt independence criterion~\cite{Wang2021}. The HSIC measures the dependence degree between finite number of observations\footnote{We here consider $\mathbf{x} = \left\{x_i \right\}_{i=1}^n$ and $\mathbf{y} = \left\{y_i \right\}_{i=1}^n$ as vectors. However, HSIC can be also used to calculate the dependence degree between matrices.} $\left(x_i, y_i \right)$, $i=1, \ldots, n$. Assume $\mathbf{K}_{\mathbf{x}}$ and $\mathbf{K}_{\mathbf{y}}$ as the kernel matrices of $\mathbf{x}$ and $\mathbf{y}$, respectively. In our analysis, we consider the radial basis function (RBF) kernel, defined by 
\begin{equation*}
    K^{RBF}(z_i, z_{i'}) = e^{-\frac{1}{n} \left(z_i-z_{i'}\right)^2},
\end{equation*}
and the linear kernel, defined by
\begin{equation*}
    K^{linear}(z_i, z_{i'}) = z_i z_{i'}.
\end{equation*}
However, other kernels can also be considered (see~\cite{Scholkopf2002} for details). An empirical calculation of HSIC is given by
\begin{equation*}
HSIC(\mathbf{x},\mathbf{y}) = \frac{\text{tr}\left(\mathbf{K}_{\mathbf{x}}\mathbf{H}\mathbf{K}_{\mathbf{y}}\mathbf{H} \right)}{\left(n-1\right)^2},
\end{equation*}
where $\mathbf{H} = \mathbf{I} - n^{-1}\mathbf{e}\mathbf{e}^T$ is the centering matrix and $\mathbf{e}$ is a $n$-dimensional column vector of ones. Note that, even if the dataset $\mathbf{x}$ is centered, this is not necessarily true for the kernel $\mathbf{K}_{\mathbf{x}}$. In order to centralize this kernel, one applies the centering matrix $\mathbf{H}$ on both sides of $\mathbf{K}_{\mathbf{x}}$. Recall that \begin{equation*}
\text{tr}\left(\mathbf{K}_{\mathbf{x}}\mathbf{H}\mathbf{K}_{\mathbf{y}}\mathbf{H} \right) = \text{tr}\left(\mathbf{H}\mathbf{K}_{\mathbf{x}}\mathbf{H}\mathbf{K}_{\mathbf{y}} \right) = \text{tr}\left(\tilde{\mathbf{K}}_{\mathbf{x}}\mathbf{K}_{\mathbf{y}} \right),
\end{equation*}
where $\tilde{\mathbf{K}}_{\mathbf{x}}$ is the centered kernel. Moreover, when we multiply $\mathbf{K}_{\mathbf{x}}$ on the left (resp. right) by $\mathbf{H}$, one removes its columns (resp. rows) mean.

As highlighted in~\cite{Li2022}, the HSIC calculation is sensitive to the scale of the observations and, therefore, an appropriate normalization should be conducted in order to compare relative dependence degrees. We thus consider a normalized version of HSIC called NOCCO (NOrmalized Cross-Covariance Operator)~\cite{Fukumizu2007}. It is defined as follows:
\begin{equation*}
NOCCO(\mathbf{x},\mathbf{y}) = \text{tr}\left(\mathbf{R}_{\mathbf{x}}\mathbf{R}_{\mathbf{y}}\right),
\end{equation*}
where 
\begin{itemize}
\item $\mathbf{R}_{\mathbf{x}} = \mathbf{H}\mathbf{K}_{\mathbf{x}}\mathbf{H} \left(\mathbf{H}\mathbf{K}_{\mathbf{x}}\mathbf{H} + n\epsilon \mathbf{I}_{n} \right)^{-1}$, 
\item $\mathbf{R}_{\mathbf{y}} = \mathbf{H}\mathbf{K}_{\mathbf{y}}\mathbf{H} \left(\mathbf{H}\mathbf{K}_{\mathbf{y}}\mathbf{H} + n\epsilon \mathbf{I}_{n} \right)^{-1}$, 
\item $\epsilon$ is a regularization parameter ({\it e.g.}, $10^{-6}$),  and 
\item $\mathbf{I}_{n}$ is a $n \times n$ identity matrix.
\end{itemize}
It is important to highlight that the use of NOCCO has two main  advantages. Firstly, as the calculation is based on the trace of matrix products, it is easily  calculated (we do not need to estimate, for instance, probability density functions). Secondly, the use of kernels brings more information than second-order moment to measure the relation between variables. Therefore, it is useful to estimate the dependence degree between them.

\section{Proposed approach}
\label{sec:proposal}

The purpose of this work is to automatically detect sensitive features and the associated impacted groups. Our hypothesis is that if a feature that provides information that splits groups of people ({\it e.g.}, if the individual is white or black) is important to increase the performance of a machine learning model (measure by means of NOCCO), then this feature may create disparities between such groups in terms of performance measures.

Consider the extended dataset $\tilde{\mathbf{X}}$, with columns $\tilde{\mathbf{X}}^{(1)}, \ldots, \tilde{\mathbf{X}}^{(m)}$, obtained after encoding categorical features. As mentioned in Section~\ref{sec:hsic}, HSIC can be used to measure the dependence between two vectors. For each numerical feature and encoded features ({\it i.e.}, for all columns of $\tilde{\mathbf{X}}$), we calculate NOCCO with respect to the vector of outcomes $\mathbf{y}$. For numerical features, the dependence degree with respect to the vector of labels is the NOCCO itself. In the case of categorical features, if this feature is a binary one, then the NOCCO for both groups will be the same and, therefore, we only need to calculate itonce. Otherwise, if the categorical features have three or more categories,  we assume as the dependence degree the maximum NOCCO among the associated encoded features. This procedure leads to $m$ measures of dependence
\begin{equation}
d_j = \text{tr}\left(\mathbf{R}_{G_j}\mathbf{R}_{\mathbf{y}}\right), \, \,\ j=1, \ldots, m,
\end{equation}
where $\mathbf{R}_{G_j} = \mathbf{H}\mathbf{K}_{G_j}\mathbf{H} \left(\mathbf{H}\mathbf{K}_{G_j}\mathbf{H} + n\epsilon \mathbf{I}_{n} \right)^{-1}$ and $\mathbf{K}_{G_j}$ is the kernel matrix associated with feature $G_j$. In the example of Section~\ref{sec:notation}, if NOCCO for $age\_\text{<30}$ is greater than the NOCCO for both $age\_\text{30-60}$ and $age\_\text{>60}$, we define $d_3$ as the dependence measure between $\tilde{\mathbf{X}}^{(4)}$ and $\mathbf{y}$.

Based on all $d_j$, $j=1, \ldots, m$, we may evaluate whether a feature should be considered as a sensitive one. In this paper, we assume an automatic procedure that takes into account a predefined threshold $t$. Algorithm~\ref{alg:proposal} presents a pseudo-code of our proposal. For each feature $G_{j}$, if $d_{j} \geq t$, $G_{j}$ is considered  sensitive  and the $G_{j,k'}$ that leads to the higher NOCCO is the harmed group. Otherwise, we do not consider $G_{j}$ as sensitive. In our empirical setting, we take $t$ to be the median value among all $d_1, \ldots, d_m$. Note that this can be easily adapted to integrate other values of $t$.

\begin{algorithm}[ht]
    \caption{(\textit{Automatic approach for sensitive features detection})}
    \label{alg:proposal}
		\begin{algorithmic}
				\STATE \textbf{Input:} $\mathbf{X}$ and $\mathbf{y}$.
				
				\STATE \textbf{Output:} Set of sensitive features $\mathcal{S} = \left\{G_{j'}, G_{j''}, \ldots \right\}$ and sensitive groups $\mathcal{S}^g = \left\{G_{j',k'}, G_{j'',k''}, \ldots \right\}$.
				
				\STATE 1: \textbf{Encoding categorical features:} Encode the categorical features and obtain the extended dataset $\tilde{\mathbf{X}} \leftarrow encode\left( \mathbf{X}\right)$.
				
				\STATE 2: \textbf{Calculate the kernel matrix of $\mathbf{y}$}: $\mathbf{K}_{\mathbf{y}} = kernel\left( \mathbf{y} \right)$
								
				\STATE 3: \textbf{Calculate the NOCCO for all features and subfeatures}:
				\FOR{$j \in \left\{1, \ldots, m \right\}$}
					\IF{$G_j$ is either a numerical or binary feature}
                        \STATE \textbf{Calculate the kernel matrix associated with $G_j$}: $\mathbf{K}_{G_j} = kernel\left( G_j \right)$
                        \STATE \textbf{Calculate the dependence measure}: $d_j = \text{tr}\left(\mathbf{R}_{G_j}\mathbf{R}_{\mathbf{y}}\right)$
                    \ELSE
                        \STATE \textbf{Calculate the maximum NOCCO for the categorical variable:}
                            \FOR{$k \in \left\{1, \ldots, q \right\}$}
                                \STATE \textbf{Calculate the kernel matrix associated with $G_{j,k}$}: $\mathbf{K}_{G_{j,k}} = kernel\left( G_{j,k} \right)$
                                \STATE \textbf{Calculate the dependence measure}: $d_j^k = \text{tr}\left(\mathbf{R}_{G_{j,k}}\mathbf{R}_{\mathbf{y}}\right)$
                            \ENDFOR
                            \STATE \textbf{Define the dependence measure for feature $G_j$}: $d_j \leftarrow \max\left(d_j^1, \ldots, d_j^q \right)$
				    \ENDIF
                \ENDFOR
                \STATE 4: \textbf{Create the set of sensitive features and sensitive groups}: $\mathcal{S} = \emptyset$ and $\mathcal{S}^g = \emptyset$
				\STATE 5: \textbf{Calculate the median value among all $d_j$, $j=1, \ldots, m$}: $t = median\left(d_1, \ldots, d_m \right)$
				\FOR{$j \in \left\{1, \ldots, m \right\}$}
                    \IF{$G_j$ is candidate for sensitive feature and $d_j \geq t$:}
                        \STATE \textbf{Update the set of sensitive features}: $\mathcal{S} \leftarrow \left\{\mathcal{S}, G_j \right\}$
                        \STATE \textbf{Update the set of sensitive groups}: $\mathcal{S}_g \leftarrow \left\{\mathcal{S}_g, G_{j,k^*} \right\}$ such that $d_j^{k^*} = \max\left(d_j^1, \ldots, d_j^q \right)$
                    \ENDIF
                \ENDFOR
    \end{algorithmic}
\end{algorithm}

\section{Empirical Evaluation}
\label{sec:exper}

In this section, we present the experimental setup to evaluate our proposal on several real datasets. We identify possible sensitive features as well as the associated sensitive groups. Moreover, we provide a comparison between the obtained NOCCO values and the fairness measures with respect to those features with high NOCCO values. 

\subsection{Datasets}

In order to assess our proposal, we consider four datasets frequently used in the literature: Adult income\footnote{\url{https://archive.ics.uci.edu/ml/datasets/adult}}, COMPAS recidivism risk~\cite{Angwin2016}, Law School Admission Council (LSAC)~\cite{Wightman1998} and Taiwanese credit default~\cite{Yeh2009}. A brief description and how we preprocessed some features are provided in the sequel. See~\cite{LeQuy2022} for further details as well as the list of features frequently considered as sensitive ones in the literature.

\begin{itemize}
    \item \textbf{Adult income dataset}: In this dataset, the aim is to predict whether a person makes over 50K a year based on the following features: age, workclass, educational-num (educational degree), marital-status, occupation, relationship (husband, not in family, other relative, own child, unmarried or wife), race (Indian-Eskimo, Asian-Pacific Islander, black, white or other), gender (male or female), capital-gain, capital-loss, hours-per-week and native-country. After removing samples with missing values, we achieved 45222 samples. We also rearranged some categorical features: age = $\left\{<\text{25, 25-60, }>\text{60}\right\}$, workclass = $\left\{\text{private, non-private}\right\}$, marital-status = $\left\{\text{married, never-married, other}\right\}$, occupation = $\left\{\text{office, heavy-work, service, other}\right\}$ and native-country = $\left\{\text{US, non-US}\right\}$. One typically assumes age, race and gender as sensitive features.

    \item \textbf{COMPAS recidivism risk dataset}: This dataset has the goal of classifying individuals as a potential criminal recidivist. There are 6167 samples and 8 features, namely sex (male or female), age\_cat (age category - less than 25, between 25 and 45 or greater than 45), race, juv\_fel\_count (number of juvenile felony), juv\_misd\_count (number of juvenile misdemeanor), juv\_other\_count (number of others infractions), priors\_count (number of priors) and c\_charge\_degree (charge degree - felony or misdemeanor). We here rearranged race $\left\{\text{Caucasian, African-American, Other}\right\}$. Race and sex are assumed as sensitive features.

    \item \textbf{Law School Admission Council (LSAC) dataset}: In this dataset, there are 23726 candidates described by 11 features: decile1b (decile given the grades in the 1st year), decile3 (decile given the grades in the 3rd year), lsat (score), ugpa (undergraduate GPA), zfygpa (1st year law school GPA), zgpa (cumulative law school GPA), fulltime (full-time or part-time work), fam\_inc (family income), male (whether the student is male or female), race and tier (tier of the law school). The goal is to predict whether a candidate would pass the bar exam. One assumes gender and race as sensitive features.
    
    \item \textbf{Taiwanese credit default dataset}: In this dataset, the aim is to predict customers' default payments in a Taiwanese institution. The 23 features are the following: limit\_bal (amount of given credit), sex (male or female), education (graduate school, university, high school or others), marriage (married, single or others), age (less than 35 or at least 35), repayment status (six features, from April to September), amount of bill statement (six features, from April to September) and amount paid (six features, from April to September). We assumed sex, education and marriage as sensitive features. 
\end{itemize}

\subsection{Fairness metrics}

In order to attest our hypothesis that the NOCCO value can be associated with fairness measures (and, hence, used as a preprocessing approach to detect sensitive features), we compare the obtained values of fairness measures with respect to features with high NOCCO values. These measures are based on the following performances extracted from the trained ML model:
\begin{itemize}
    \item True positive ($TP$): \# of instances correctly classified as class 1.
    \item True negative ($TN$): \#  of instances correctly classified as class -1.
    \item False positive ($FP$): \#  of instances wrongly classified as class 1.
    \item False positive ($FN$): \#  of instances wrongly classified as class -1.    
\end{itemize}
Aiming at evaluating some fairness measures when splitting the dataset into different groups of people, we restrict the aforementioned performance measures to such groups. Suppose, for example, that feature $G_{j}$ contains information that splits people into groups $\left\{G_{j,1}, \ldots, G_{j,q} \right\}$. As the NOCCO for group $G_{j,k}$ will indicate the dependence degree between such a group and the vector of labels, when evaluating fairness, we consider the disparities when people is divided between those belonging to group $G_{j,k}$ and those that does not. Therefore, we restrict the performance measures such that $FP_{G_{j,k}}$ means the number of instances belonging to group $G_{j,k}$ wrongly classified as class 1 and $TN_{-G_{j,k}}$ is the number of instances not belonging to group $G_{j,k}$ correctly classified as class -1 (we referred to $-G_{j,k}$ as the instances not belonging to group $G_{j,k}$).

Given the performance measures conditioned on groups, we may define the fairness measures considered in our experiments:
\begin{itemize}
    \item Predictive equality ($PE$): A classifier satisfies predictive equality if both groups have equal false positive rate. In other words, we should have low vales of
\begin{equation*}
    f_{PE} = \left| \frac{FP_{G_{j,k}}}{FP_{G_{j,k}} + TN_{G_{j,k}}} - \frac{FP_{-G_{j,k}}}{FP_{-G_{j,k}} + TN_{-G_{j,k}}} \right|.
\end{equation*}
    \item Equal opportunity ($EP$): A classifier satisfies equal opportunity if both groups have equal true positive rate. In other words, we should have low vales of
\begin{equation*}
    f_{EP} = \left| \frac{TP_{G_{j,k}}}{TP_{G_{j,k}} + FN_{G_{j,k}}} - \frac{TP_{-G_{j,k}}}{TP_{-G_{j,k}} + FN_{-G_{j,k}}} \right|.
\end{equation*}
    \item Equalized odds ($EO$): A classifier satisfies equalized odds if both groups have equal true positive and false positive rates. In other words, we should have low vales of
\begin{equation*}
    f_{EO} =  \left| \frac{TP_{G_{j,k}}}{TP_{G_{j,k}} + FN_{G_{j,k}}} - \frac{TP_{-G_{j,k}}}{TP_{-G_{j,k}} + FN_{-G_{j,k}}} \right| 
     + \left| \frac{FP_{G_{j,k}}}{FP_{G_{j,k}} + TN_{G_{j,k}}} - \frac{FP_{-G_{j,k}}}{FP_{-G_{j,k}} + TN_{-G_{j,k}}} \right|.
\end{equation*}
    \item Overall accuracy equality ($OAE$): A classifier satisfies overall accuracy equality if both groups have equal correct classification for both classes -1 and 1. In other words, we should have low vales of
\begin{equation*}
    f_{OAE} = \left| \frac{TP_{G_{j,k}} + TN_{G_{j,k}}}{n_{G_{j,k}}} - \frac{TP_{-G_{j,k}} + TN_{-G_{j,k}}}{n_{-G_{j,k}}} \right|,
\end{equation*}
where $n_{G_{j,k}}$ and $n_{-G_{j,k}}$ are the number of instances that belong and do not belong to group $G_{j,k}$.
\end{itemize}

\subsection{Result analysis}

In our experiments, we first calculated the NOCCO for each feature and encoded feature. Thereafter, we calculate the median NOCCO and define the features that could be considered as sensitive ones. Aiming at attesting the usefulness of NOCCO as a measure to detect sensitive features, we comparing the NOCCO values with a set of fairness measures. For this purpose, we trained a Random Forest classifier using k-fold cross validation (with 10 folds). For each fold and encoded features, we calculated performances and fairness measures (equal opportunity, predictive parity, equalized odds and overall accuracy equality, by taking the absolute difference). For categorical features with more than two groups, we calculated these measures by assuming a one-vs-all strategy.

\begin{figure}[!h]
\centering
\subfloat[Dependence measure (NOCCO).]{\includegraphics[width=3.3in]{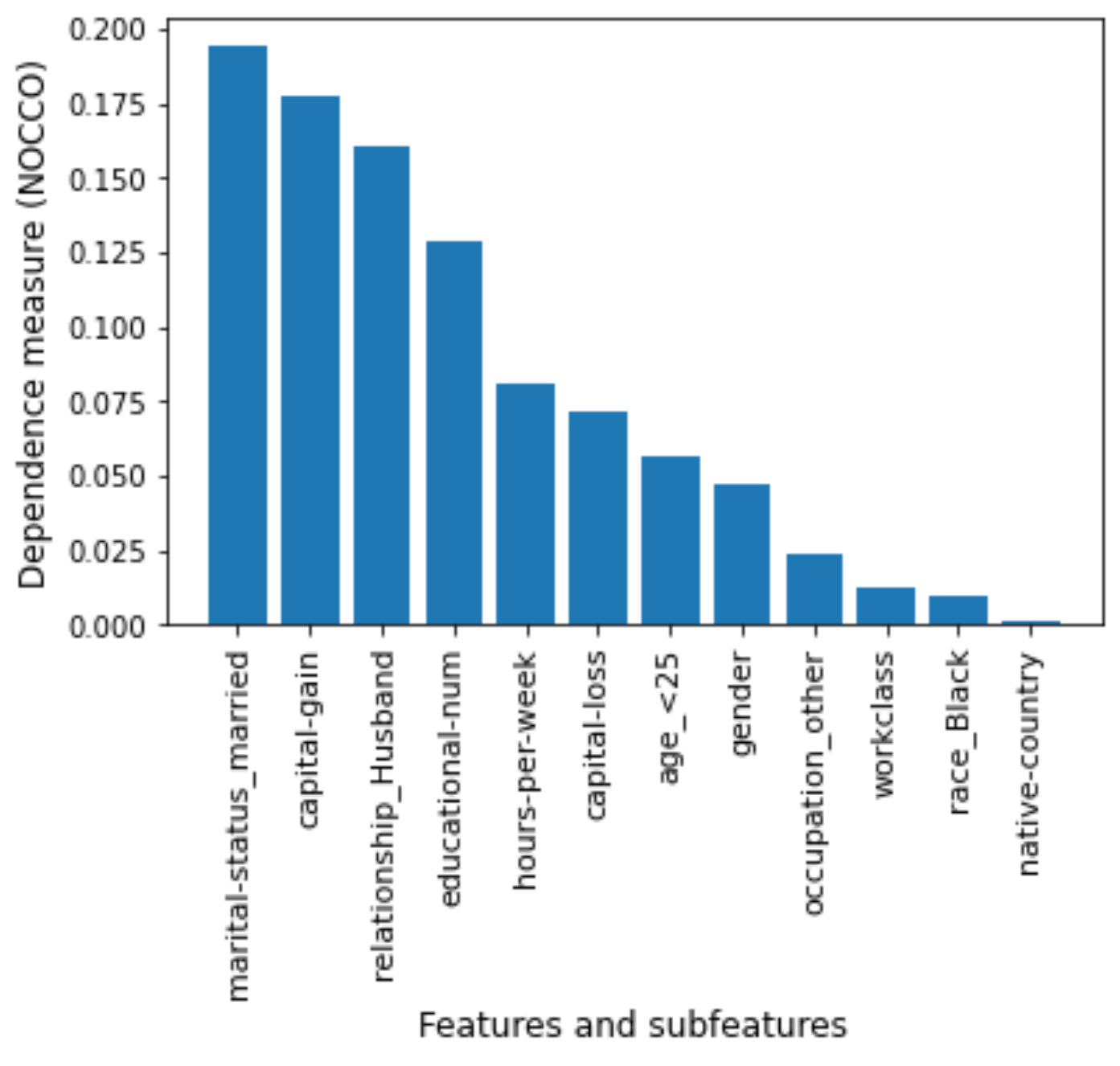}
\label{fig:adult_nocco_RBF}}
\hfill
\subfloat[Relation between NOCCO and fairness measures.]{\includegraphics[width=4.3in]{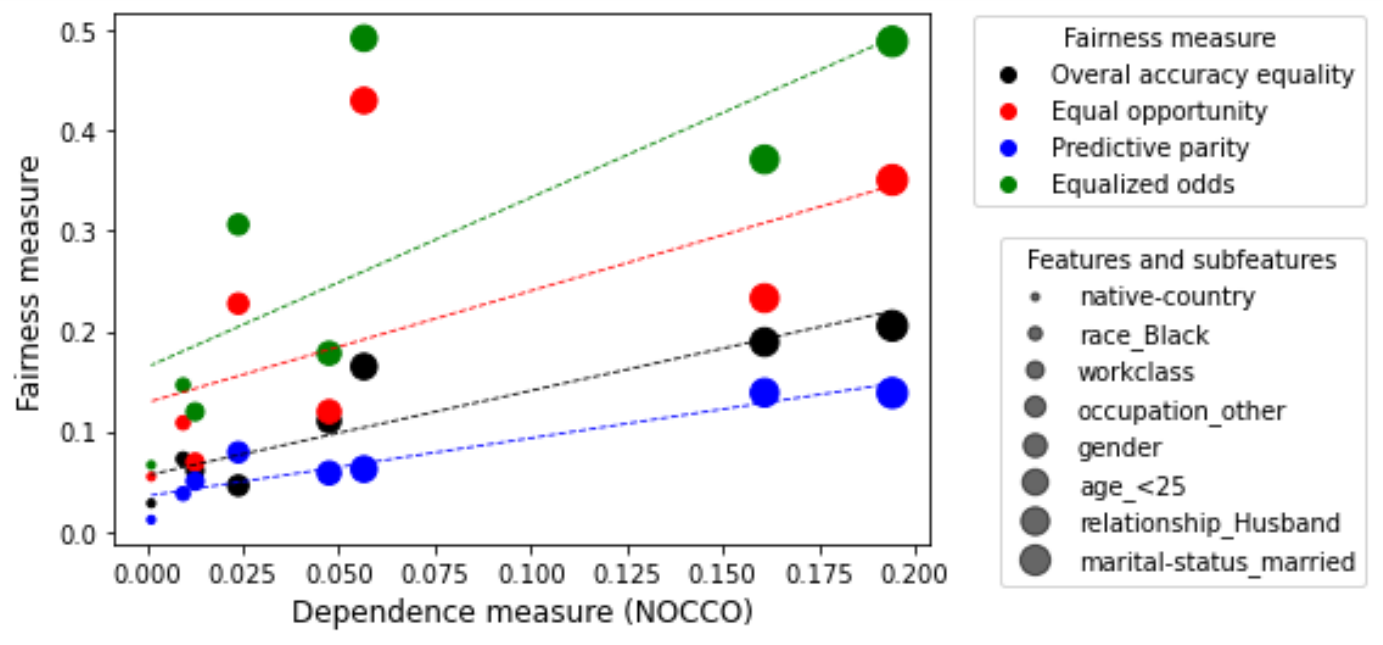}
\label{fig:adult_perf_RBF}}
\hfill
\caption{Results for the Adult income dataset (with RBF kernel).}
\label{fig:adult_RBF}
\end{figure}

\begin{figure}[!h]
\centering
\subfloat[Dependence measure (NOCCO).]{\includegraphics[width=3.0in]{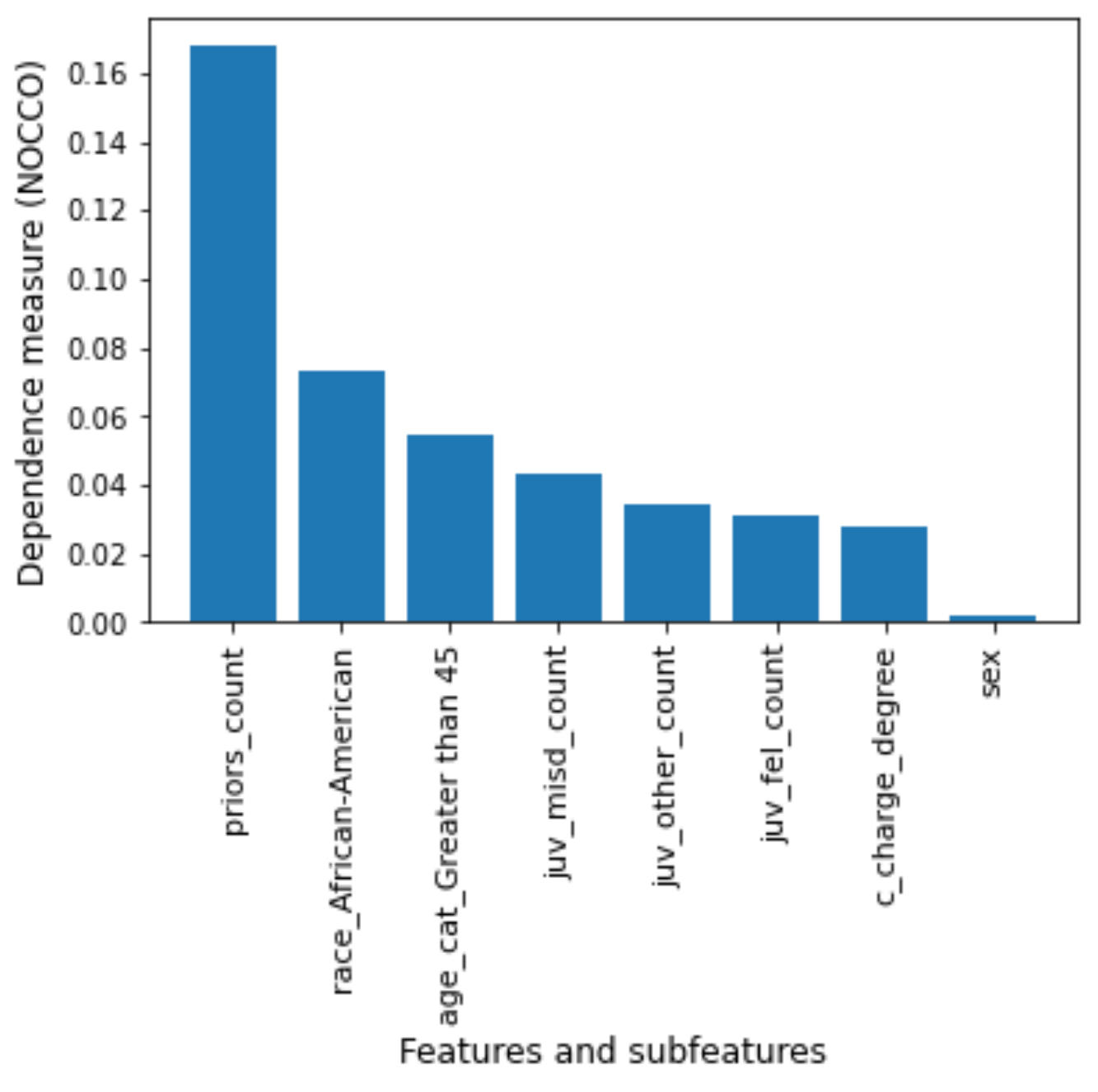}
\label{fig:compas_nocco_RBF}}
\hfill
\subfloat[Relation between NOCCO and fairness measures.]{\includegraphics[width=4.0in]{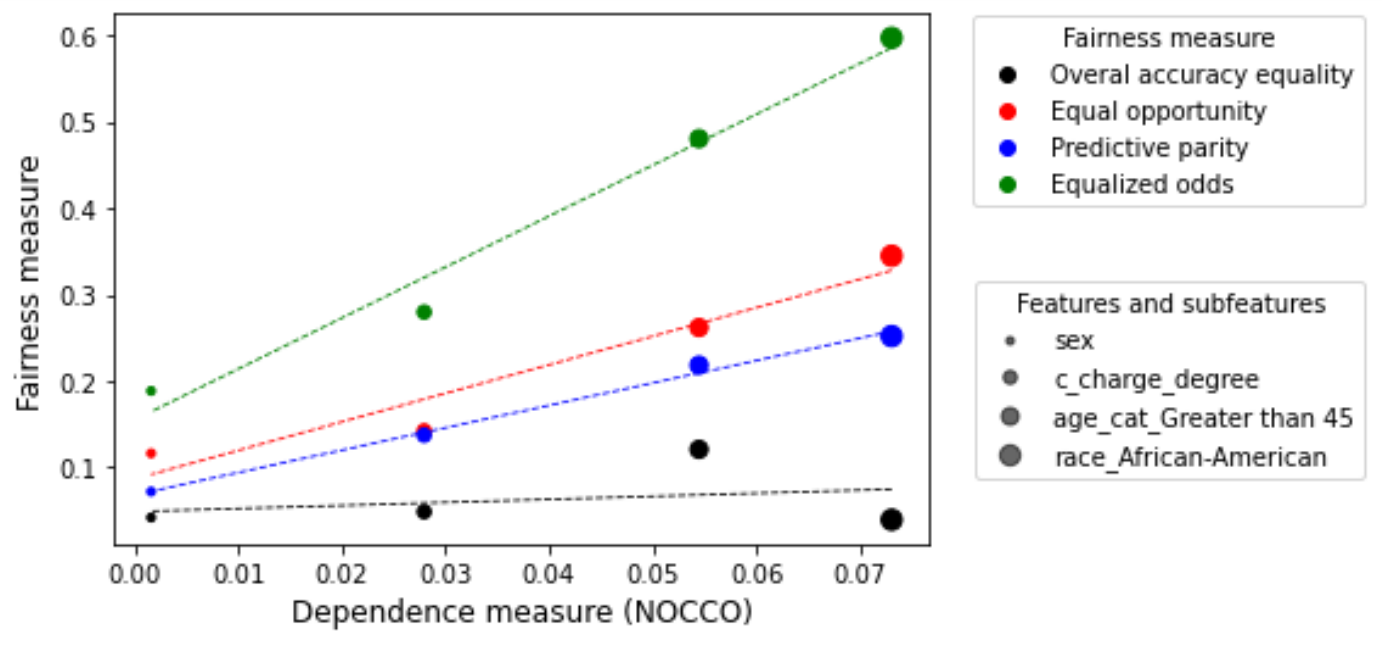}
\label{fig:compas_perf_RBF}}
\hfill
\caption{Results for the COMPAS dataset (with RBF kernel).}
\label{fig:compas_RBF}
\end{figure}

\begin{figure}[!h]
\centering
\subfloat[Dependence measure (NOCCO).]{\includegraphics[width=3in]{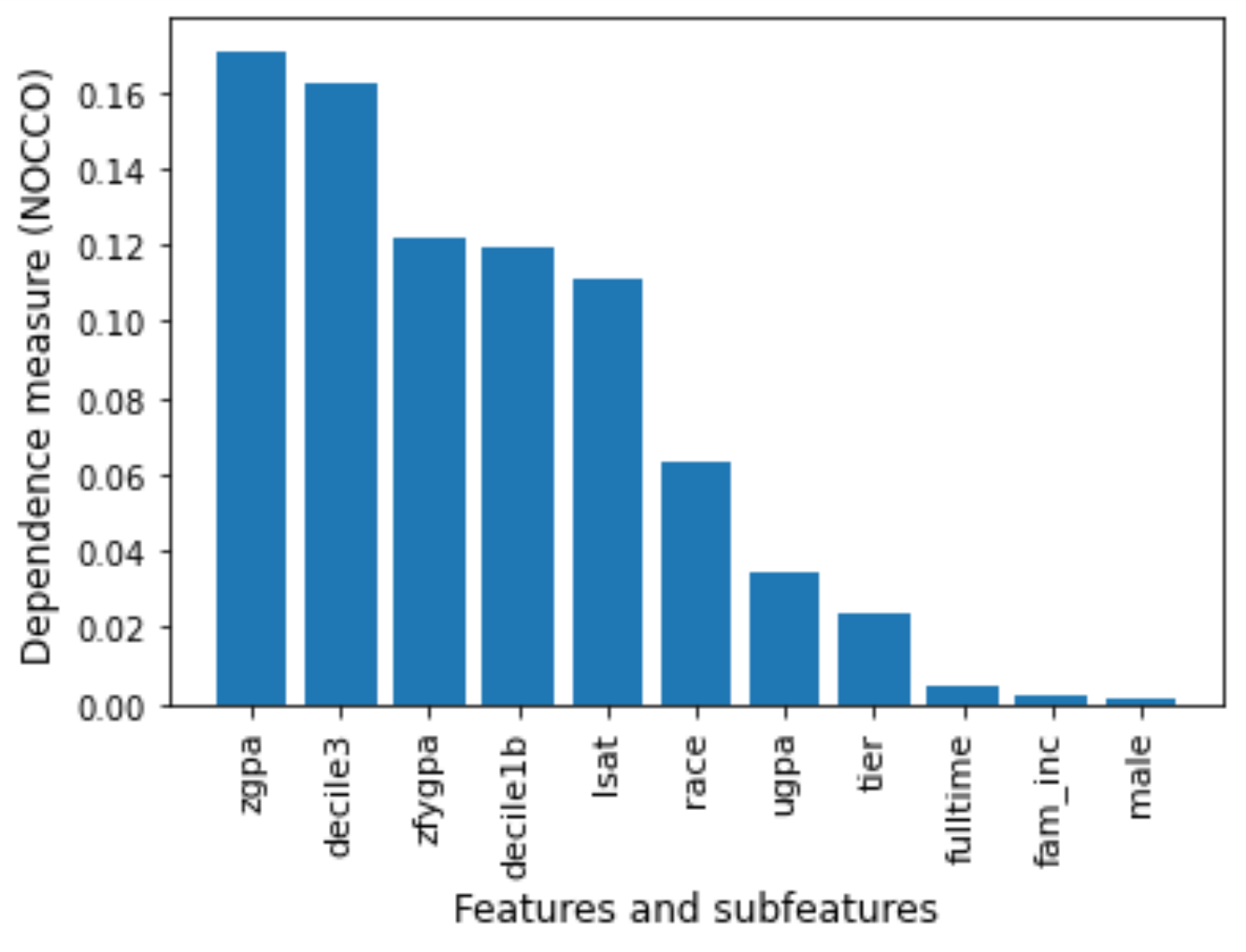}
\label{fig:lsac_nocco_RBF}}
\hfill
\subfloat[Relation between NOCCO and fairness measures.]{\includegraphics[width=4.0in]{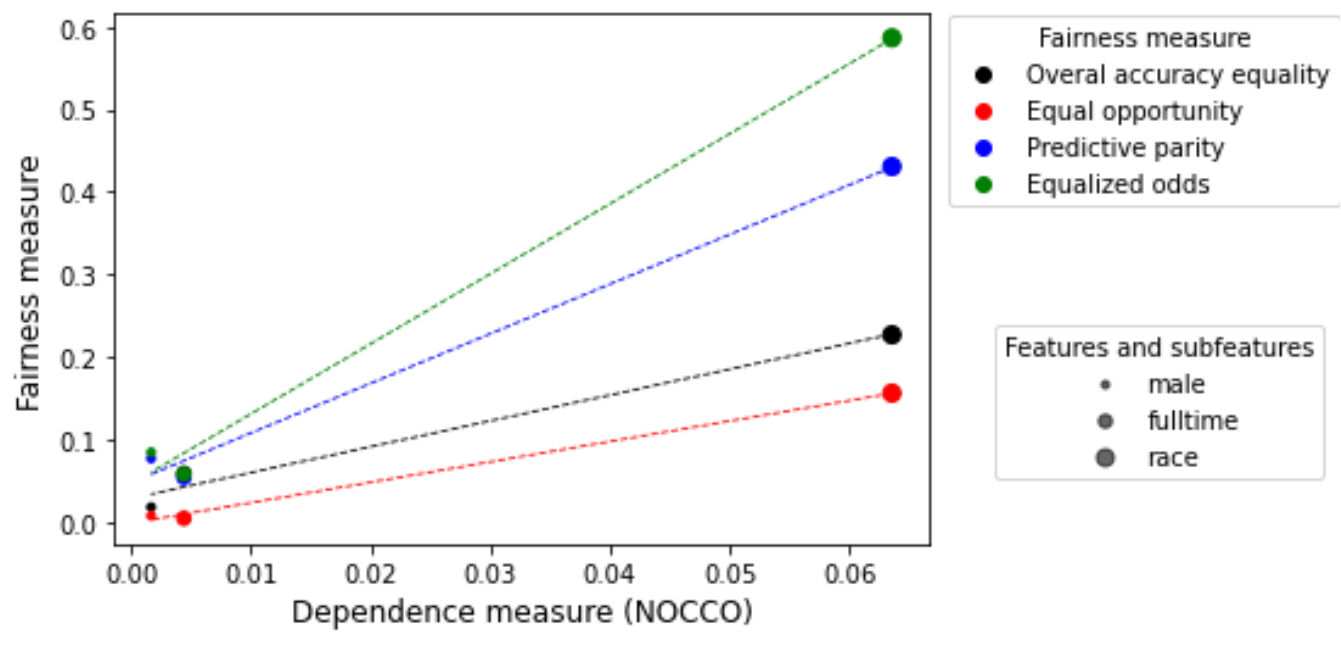}
\label{fig:lsac_perf_RBF}}
\hfill
\caption{Results for the LSAC dataset (with RBF kernel).}
\label{fig:lsac_RBF}
\end{figure}

\begin{figure}[!h]
\centering
\subfloat[Dependence measure (NOCCO).]{\includegraphics[width=3.3in]{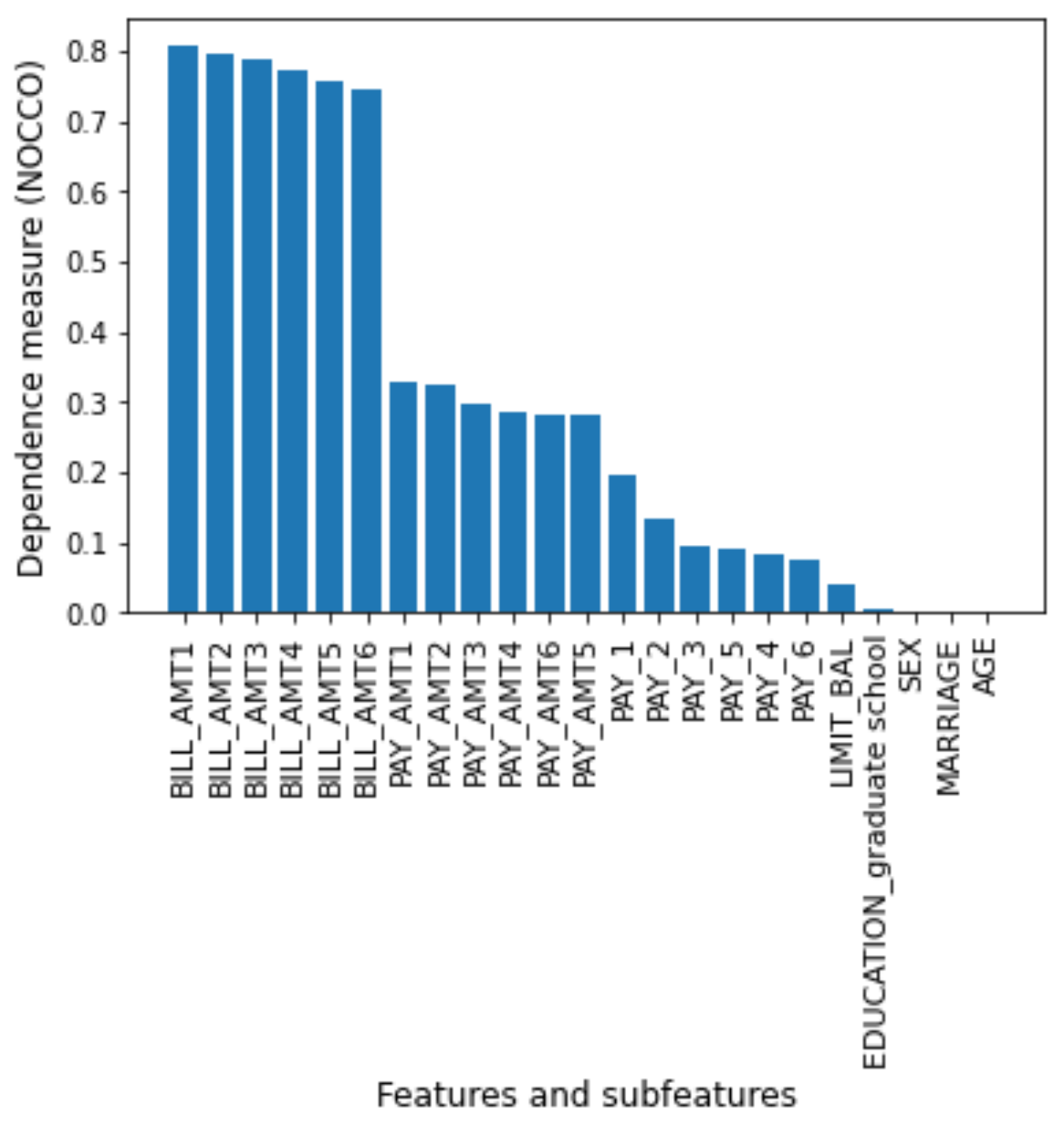}
\label{fig:tcred_nocco_RBF}}
\hfill
\subfloat[Relation between NOCCO and fairness measures.]{\includegraphics[width=4.3in]{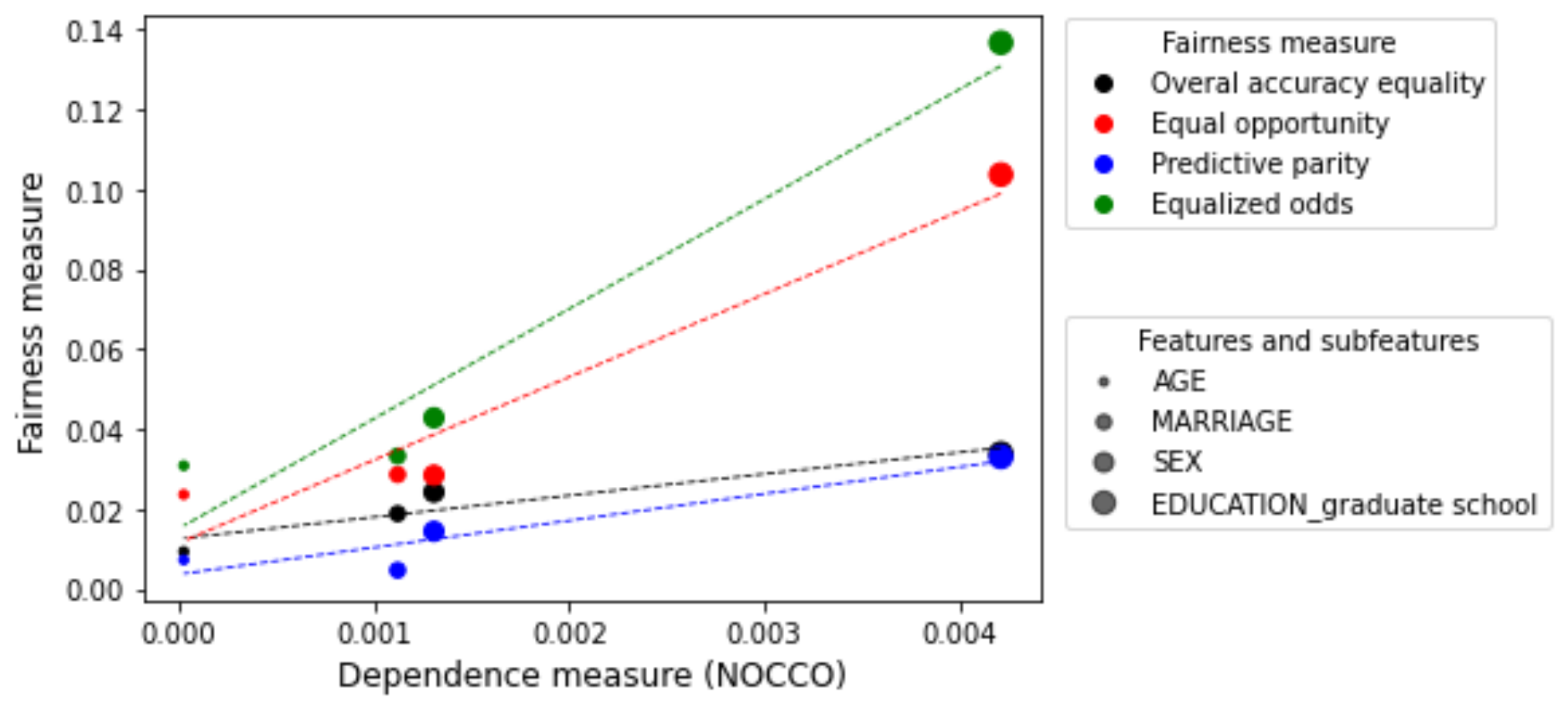}
\label{fig:tcred_perf_RBF}}
\hfill
\caption{Results for the Taiwanese credit default dataset (with RBF kernel).}
\label{fig:tcred_RBF}
\end{figure}

We present the obtained results for the Adult income, COMPAS, LSAC and Taiwanese credit default datasets in Figures~\ref{fig:adult_RBF},~\ref{fig:compas_RBF},~\ref{fig:lsac_RBF} and~\ref{fig:tcred_RBF}, respectively. In NOCCO calculation, we assumed the RBF kernel. Note that, for all datasets and all fairness measures (see Figures~\ref{fig:adult_perf_RBF},~\ref{fig:compas_perf_RBF},~\ref{fig:lsac_perf_RBF} and~\ref{fig:tcred_perf_RBF}), the higher is the NOCCO, the higher is the fairness measures ({\it i.e.}, the higher are the unfair results).

\paragraph{Results on the Adult income dataset} In the results with the Adult income dataset, we obtained a median NOCCO equals to 0.0642. Features whose NOCCO is greater than this threshold are marital-status, capital-gain, relationship, educational-num, hours-per-week and capital-loss (see Figure~\ref{fig:adult_nocco_RBF}). We consider that splitting people based on either education-num or hours-per-week do not create unfair concerns. However, the other two features can be seen as sensitive ones. Although age and gender are frequently pointed as sensitive features in this dataset, marital-status and relationship appeared as higher candidates. The higher NOCCO was obtained when splitting people between married and non-married. Moreover, NOCCO is high when we divide the instances in those who are a husband in a relation and those who are not. As in this dataset, husband in a relation is a proxy to know that the person is a man, this information is directly associated with gender. Therefore, not only gender should be consider as a sensitive feature when looking at men/women, but we should also be aware of the information provided by relationship. Another interesting aspect in this dataset is that race led to a very small NOCCO. This suggested that, although frequently considered as a sensitive feature, in this problem, it will not lead to disparate results. We attested this conclusion in the fairness measures presented in Figure~\ref{fig:adult_perf_RBF}. The fairness measures are very small when splitting people by race.

\paragraph{Results on the COMPAS dataset} In COMPAS dataset, the literature assumes gender and race as sensitive features. In the NOCCO values presented in Figure~\ref{fig:compas_nocco_RBF} and with $t = 0.0390$, only race (when dividing people between African-American and other ones) and age category (greater than 45) could be considered as sensitive features. Although sex is frequently considered as sensitive, the obtained NOCCO value indicates that, in this dataset, it does not entail disparities between male and female. See in Figure~\ref{fig:compas_perf_RBF} that fairness measures when splitting people by sex are much lower in comparison with race or age category. Another interesting remark presented in Figure~\ref{fig:compas_perf_RBF} is that the the relation between NOCCO and most of the fairness measures are nearly linear.

\paragraph{Results on the LSAC dataset}
The NOCCO values for the LSAC dataset are presented in Figure~\ref{fig:lsac_nocco_RBF}. Only race and sex is considered as candidates for sensitive features in this dataset. With a median value of 0.0635, race should be considered as a sensitive feature and male should not. Indeed, as can be attested in Figure~\ref{fig:lsac_perf_RBF}, while splitting people by race creates unfair results (we obtained high values of fairness measures), a division by gender does not (the fairness measures are very small). 

\paragraph{Results the Taiwanese credit default dataset}
Another interesting result was obtained with the Taiwanese credit default dataset. One may see in Figure~\ref{fig:tcred_nocco_RBF} that features generally considered as sensitive ones (sex, education and marriage) have very small NOCCO values. Features associated with payment status have a much higher dependence with class labels than the considered sensitive features. Therefore, with a median NOCCO value of 0.2789, no feature could be defined as sensitive. This result is attested by the small values (with most of them lower than 0.10) of the fairness metrics presented in Figure~\ref{fig:tcred_RBF}. Indeed, if one compares these fairness metrics with the ones achieved with the previous datasets, one may assume that they are small.

\subsection{Consistency with linear kernel}

\begin{figure*}[h!]
\centering
\subfloat[Adult income.]{\includegraphics[width=3.1in]{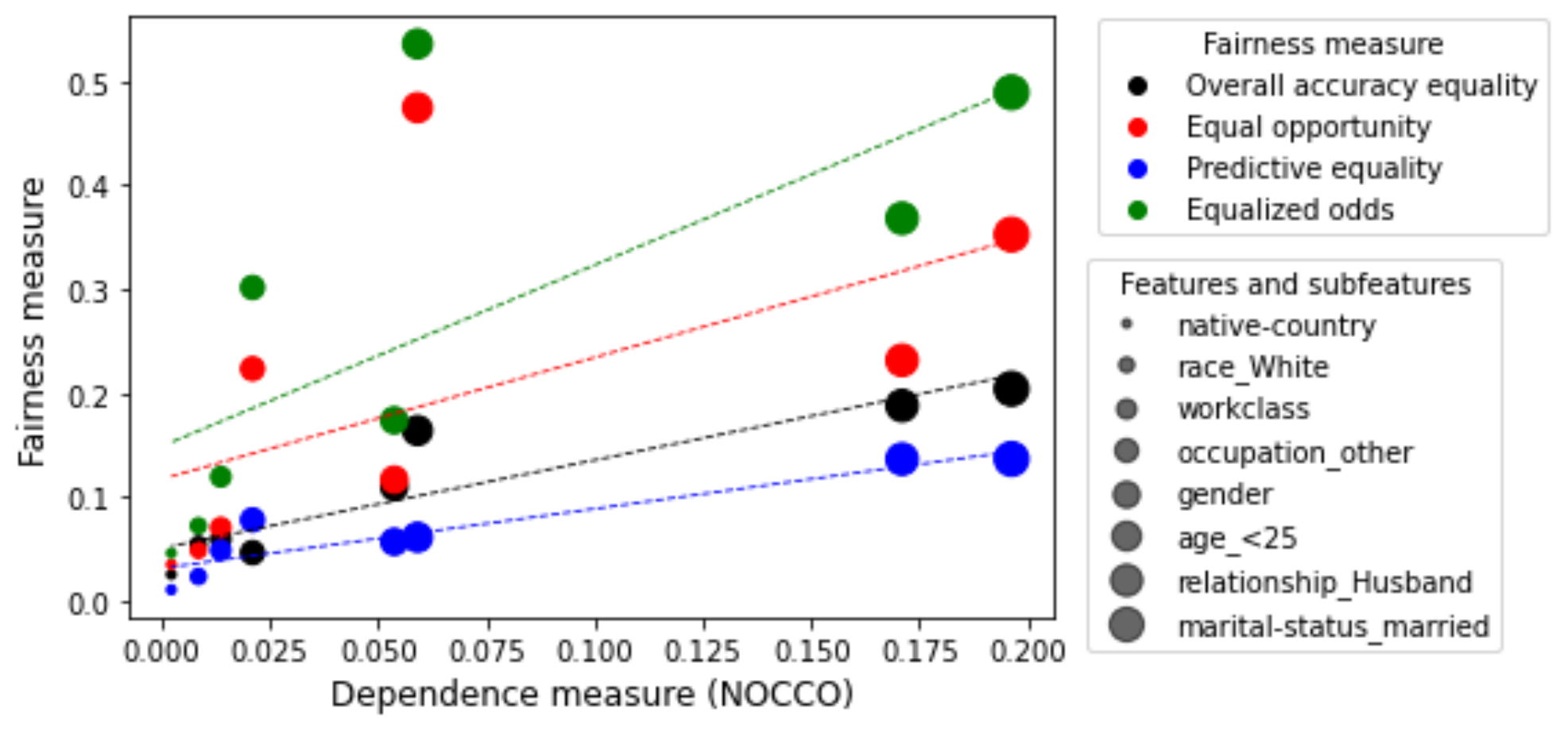}
\label{fig:adult_perf}}
\hfill
\subfloat[COMPAS.]{\includegraphics[width=3.1in]{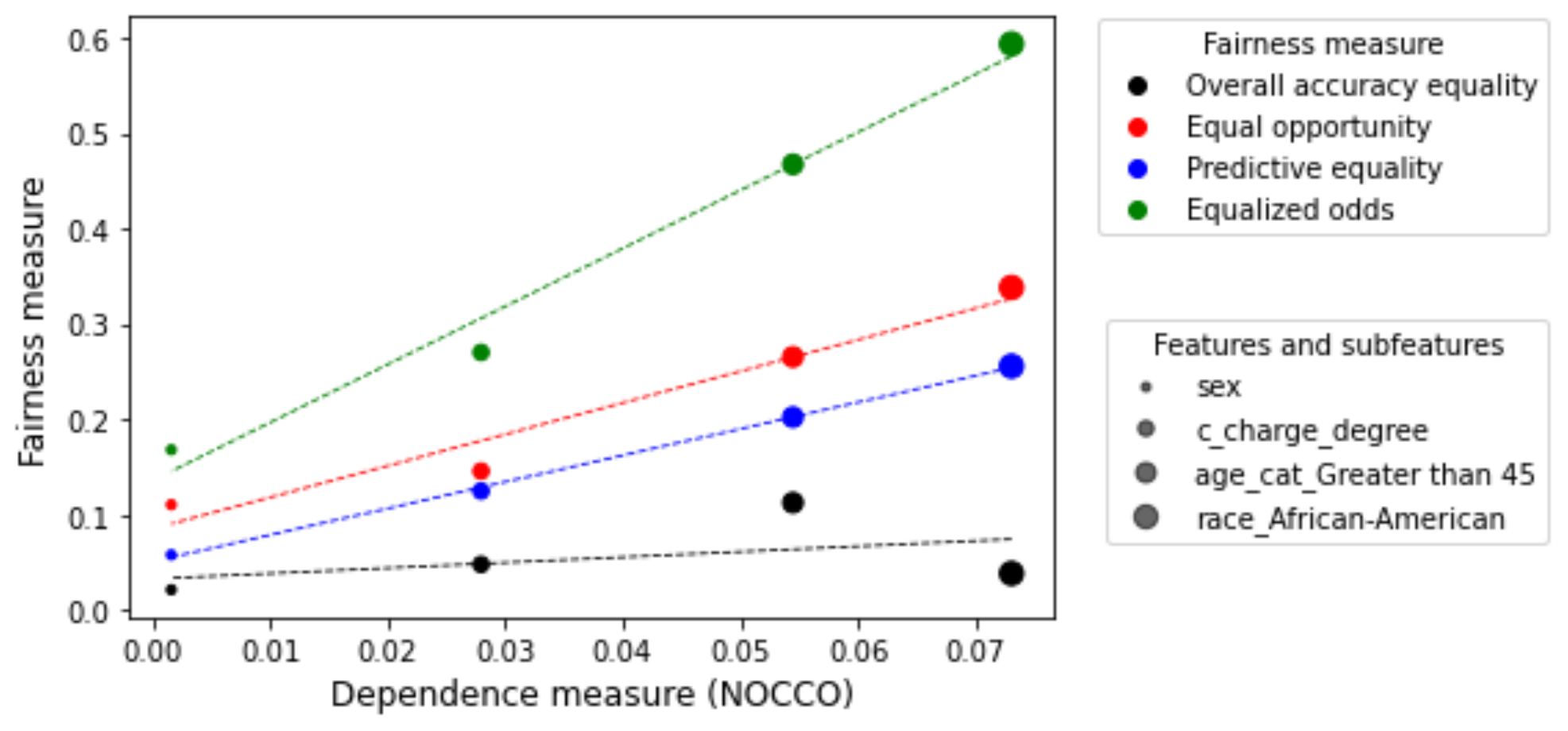}
\label{fig:compas_perf}}
\hfill
\subfloat[LSAC.]{\includegraphics[width=3.1in]{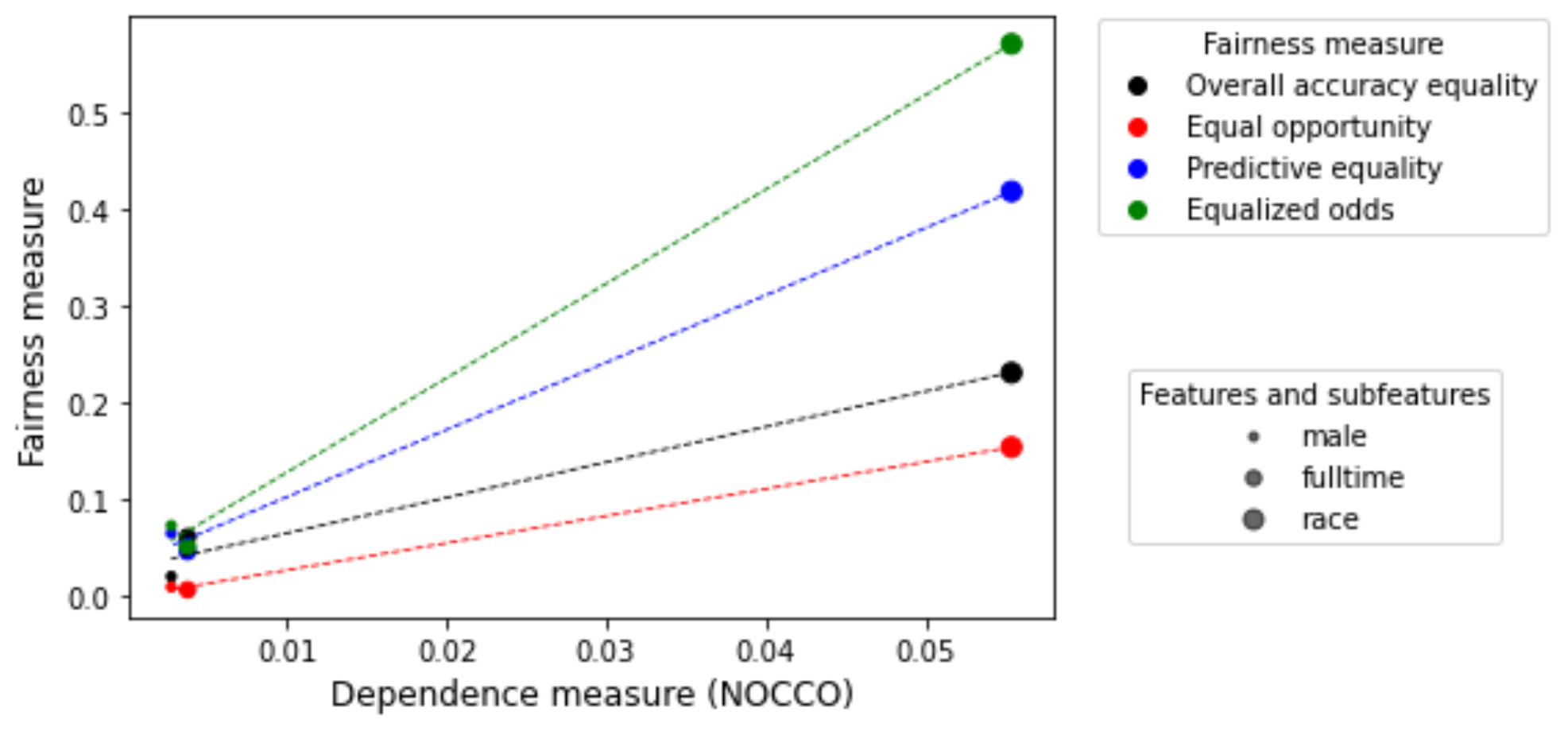}
\label{fig:lsac_perf}}
\hfill
\subfloat[Taiwanese credit default.]{\includegraphics[width=3.1in]{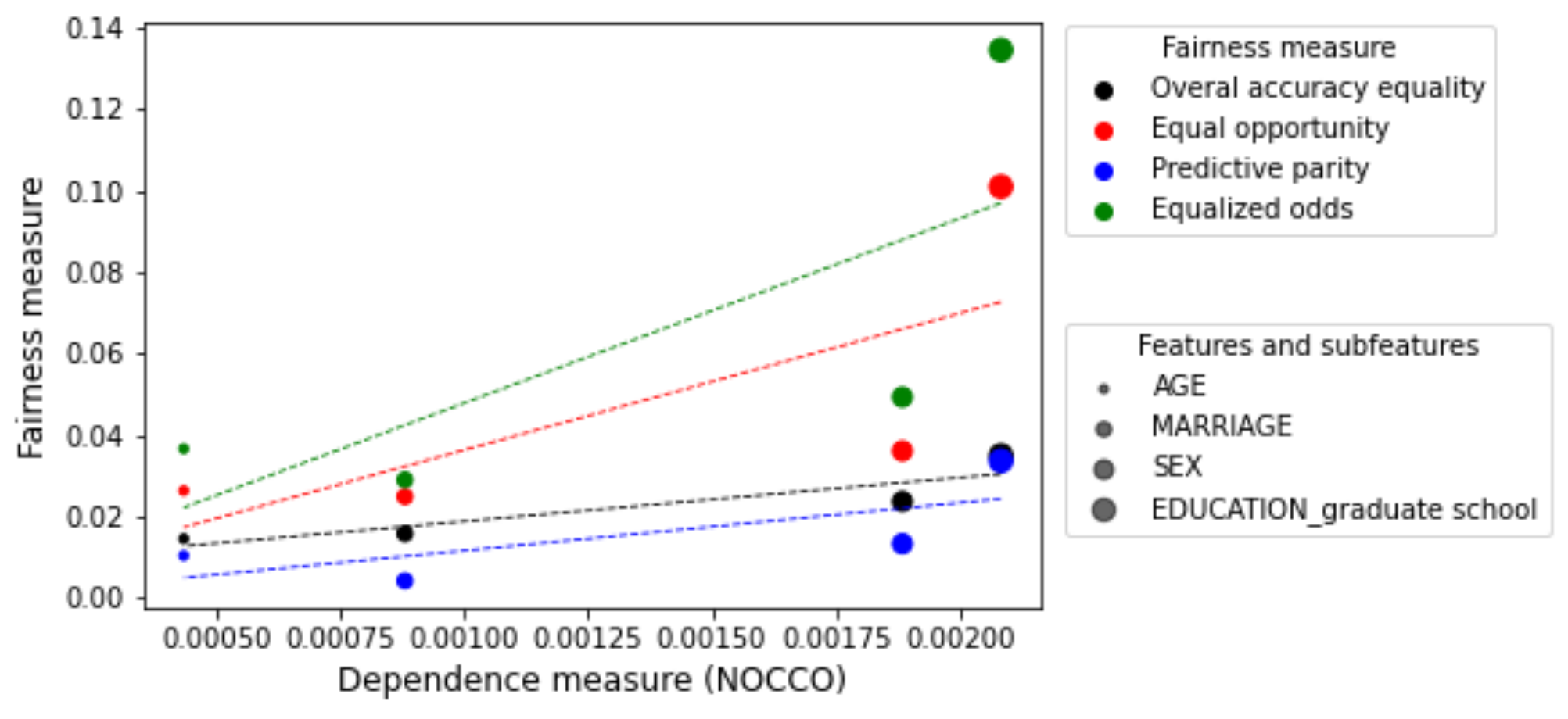}
\label{fig:tcred_perf}}
\hfill
\caption{Relation between NOCCO (with linear kernel) and fairness measures for all datasets. We compare~\ref{fig:adult_perf} (Adult income),~\ref{fig:compas_perf} (COMPAS),~\ref{fig:lsac_perf} (LSAC),~\ref{fig:tcred_perf} (Taiwanese credit default) with Figures~\ref{fig:adult_perf_RBF},~\ref{fig:compas_perf_RBF},~\ref{fig:lsac_perf_RBF} and~\ref{fig:tcred_perf_RBF}, respectively.}
\label{fig:RBF_kernel}
\end{figure*}

In order to verify the consistency of the previous results by assuming another kernel, we evaluate the relation between linear kernal-based NOCCO and the fairness measures. The results for the Adult income, COMPAS, LSAC and Taiwanese credit default datasets are presented in Figures~\ref{fig:adult_perf},~\ref{fig:compas_perf},~\ref{fig:lsac_perf} and~\ref{fig:tcred_perf}, respectively. Clearly, all the results with the linear kernel are consistent with the ones obtained by using the RBF kernel. The only difference can be seen in Adult income dataset with respect to the detected sensitive group for race. Instead of race\_White (as detected by using the RBF kernel), we here assigned race\_Black as the sensitive group. The reason for this disparity lies in the number of samples for each race category. As the number of samples whose categories are different from white or black is less than 5\%, when using one-vs-all strategy, the results for race\_White and race\_Black are practically the same. Therefore, we can safely consider that there was not a disparity between the obtained results.

\section{Conclusions and perspectives}
\label{sec:concl}

In this paper, we proposed an automatic approach to detect sensitive features based on a normalized version of the Hilbert–Schmidt independence criterion. As our method only requires the dataset and labels, we are able to detect sensitive features without the need of training a machine learning model and evaluating the performance measures. The numerical experiments on several datasets and different kernels attested that the higher is the dependence between a sensitive feature and the vector of labels, the higher are the outcome disparities with respect to groups discriminated by such a feature. Therefore, the statistical dependence calculated from the dataset and labels  proved to be a measure that can guide researchers and practitioners in detecting sensitive features, especially in situations and case scenarios where the pertaining fairness notions seem hard to identify.

We highlight that an automatic procedure to detect sensitive features is of importance in practical problems. Generally, one subjectively defines which features are sensitive. However, we showed that features frequently considered as sensitive in the literature may not entail disparate results. This finding can save effort when developing a machine learning model that takes into account information from predetermined sensitive features in order to mitigate disparate results provided by them. 

This work opens interesting perspectives for future works. Although we used the NOCCO to calculate the dependence degree between labels and vectors of features that describe groups in the population ({\it e.g.}, if the person is black or not), this measure can also be applied to coalitions of features ({\it e.g.}, subgroups). In this case, we aim at verifying the dependence degree between tow or more sensitive information ({\it e.g.}, if the person is a black woman with less than 30 years old) and the labels. This may show that combining information provided by two or more features may increase the chance of achieving disparate results. As another perspective, we intend to extend the proposed approach to deal with multiclass classification problems. In such scenarios, for each available class, we may detect which features should be considered as sensitive. \\


\textbf{Acknowledgements} \\

Work supported by S\~{a}o Paulo Research Foundation (FAPESP) under the grants \#2020/09838-0 (BI0S - Brazilian Institute of Data Science), \#2020/10572-5 and \#2021/11086-0. The research of the second named author was partially supported by TAILOR, a project funded by EU Horizon 2020 research and innovation programme under GA No 952215, and the Inria Project Lab ``Hybrid Approaches for Interpretable AI'' (HyAIAI).

\bibliographystyle{named}
\bibliography{references_ecai2023}

\begin{thebibliography}{}

\bibitem[\protect\citeauthoryear{Agarwal \bgroup \em et al.\egroup
  }{2018}]{agarwal2018reductions}
Alekh Agarwal, Alina Beygelzimer, Miroslav Dud{\'\i}k, John Langford, and Hanna
  Wallach.
\newblock A reductions approach to fair classification.
\newblock In {\em International Conference on Machine Learning}, pages 60--69.
  PMLR, 2018.

\bibitem[\protect\citeauthoryear{Alves \bgroup \em et al.\egroup
  }{2021}]{alves2021reducing}
Guilherme Alves, Maxime Amblard, Fabien Bernier, Miguel Couceiro, and Amedeo
  Napoli.
\newblock Reducing unintended bias of {ML} models on tabular and textual data.
\newblock In {\em 8th {IEEE} International Conference on Data Science and
  Advanced Analytics, {DSAA} 2021}, pages 1--10. {IEEE}, 2021.

\bibitem[\protect\citeauthoryear{Angwin \bgroup \em et al.\egroup
  }{2016}]{Angwin2016}
Julia. Angwin, Jeff. Larson, Surya. Mattu, and Lauren. Kirchner.
\newblock {Machine Bias - {P}ro{P}ublica}, 2016.

\bibitem[\protect\citeauthoryear{Barshan \bgroup \em et al.\egroup
  }{2011}]{Barshan2011}
Elnaz Barshan, Ali Ghodsi, Zohreh Azimifar, and Mansoor~Zolghadri Jahromi.
\newblock Supervised principal component analysis: Visualization,
  classification and regression on subspaces and submanifolds.
\newblock {\em Pattern Recognition}, 44:1357--1371, 2011.

\bibitem[\protect\citeauthoryear{Bhargava \bgroup \em et al.\egroup
  }{2020}]{BhargavaCN20}
Vaishnavi Bhargava, Miguel Couceiro, and Amedeo Napoli.
\newblock Limeout: An ensemble approach to improve process fairness.
\newblock In {\em {ECML} {PKDD} 2020 Workshops - Workshops of the European
  Conference on Machine Learning and Knowledge Discovery in Databases {(ECML}
  {PKDD} 2020): {XKDD} 2020 Proceedings}, volume 1323 of {\em Communications in
  Computer and Information Science}, pages 475--491. Springer, 2020.

\bibitem[\protect\citeauthoryear{Biega \bgroup \em et al.\egroup
  }{2018}]{Gummadi18}
Asia~J. Biega, Krishna~P. Gummadi, and Gerhard Weikum.
\newblock Equity of attention: Amortizing individual fairness in rankings.
\newblock In {\em The 41st International {ACM} {SIGIR} Conference on Research
  {\&} Development in Information Retrieval, {SIGIR} 2018}, pages 405--414.
  {ACM}, 2018.

\bibitem[\protect\citeauthoryear{Brink \bgroup \em et al.\egroup
  }{2016}]{Brink2016}
Henrik Brink, Joseph Richards, and Mark Fetherolf.
\newblock {\em {Real-world machine learning}}.
\newblock Simon and Schuster, 2016.

\bibitem[\protect\citeauthoryear{Calders \bgroup \em et al.\egroup
  }{2009}]{calders2009building}
Toon Calders, Faisal Kamiran, and Mykola Pechenizkiy.
\newblock Building classifiers with independency constraints.
\newblock In {\em 2009 IEEE International Conference on Data Mining Workshops},
  pages 13--18. IEEE, 2009.

\bibitem[\protect\citeauthoryear{Davidson \bgroup \em et al.\egroup
  }{2019}]{davidson2019racial}
Thomas Davidson, Debasmita Bhattacharya, and Ingmar Weber.
\newblock Racial bias in hate speech and abusive language detection datasets.
\newblock In {\em Proceedings of the Third Workshop on Abusive Language
  Online}, pages 25--35, 2019.

\bibitem[\protect\citeauthoryear{Dwork \bgroup \em et al.\egroup
  }{2012}]{dwork2012fairness}
Cynthia Dwork, Moritz Hardt, Toniann Pitassi, Omer Reingold, and Richard Zemel.
\newblock Fairness through awareness.
\newblock In {\em Proceedings of the 3rd innovations in theoretical computer
  science conference}, pages 214--226, 2012.

\bibitem[\protect\citeauthoryear{Fish \bgroup \em et al.\egroup
  }{2016}]{fish2016confidence}
Benjamin Fish, Jeremy Kun, and {\'A}d{\'a}m~D Lelkes.
\newblock A confidence-based approach for balancing fairness and accuracy.
\newblock In {\em Proceedings of the 2016 SIAM international conference on data
  mining}, pages 144--152. SIAM, 2016.

\bibitem[\protect\citeauthoryear{Fukumizu \bgroup \em et al.\egroup
  }{2007}]{Fukumizu2007}
Kenji Fukumizu, Arthur Gretton, Xiaohai Sun, and Bernhard Sch{\"{o}}lkopf.
\newblock Kernel measures of conditional dependence.
\newblock In {\em Advances in Neural Information Processing Systems 20 (NIPS)},
  2007.

\bibitem[\protect\citeauthoryear{Greenfeld and Shalit}{2020}]{Greenfeld2020}
Daniel Greenfeld and Uri Shalit.
\newblock Robust learning with the hilbert-schmidt independence criterion.
\newblock In {\em Proceedings of the 37th International Conference on Machine
  Learning}, pages 3759--3768. PMLR, 2020.

\bibitem[\protect\citeauthoryear{Gretton \bgroup \em et al.\egroup
  }{2005}]{Gretton2005}
Arthur Gretton, Olivier Bousquet, Alexander~J. Smola, and Bernhard
  Sch{\"{o}}lkopf.
\newblock Measuring statistical dependence with hilbert-schmidt norms.
\newblock In {\em Algorithmic Learning Theory, 16th International Conference,
  {ALT} 2005}, volume 3734 of {\em Lecture Notes in Computer Science}, pages
  63--77. Springer, 2005.

\bibitem[\protect\citeauthoryear{Grgić-Hlača \bgroup \em et al.\egroup
  }{2018}]{GummadiProcess18}
Nina Grgić-Hlača, Muhammad~Bilal Zafar, Krishna~P. Gummadi, and Adrian
  Weller.
\newblock Beyond distributive fairness in algorithmic decision making: Feature
  selection for procedurally fair learning.
\newblock In {\em Proceedings of the AAAI Conference on Artificial
  Intelligence}, volume~32, pages 51--60. {AAAI} Press, 2018.

\bibitem[\protect\citeauthoryear{Grgic-Hlaca \bgroup \em et al.\egroup
  }{2016}]{Gummadi2016}
Nina Grgic-Hlaca, Muhammad~Bilal Zafar, Krishna~P. Gummadi, and Adrian Weller.
\newblock The case for process fairness in learning: Feature selection for fair
  decision making.
\newblock 2016.

\bibitem[\protect\citeauthoryear{Hall}{1999}]{Hall1999}
Mark~A. Hall.
\newblock {\em {Correlation-based feature selection for machine learning}}.
\newblock PhD thesis, The University of Waikato, 1999.

\bibitem[\protect\citeauthoryear{Hangartner \bgroup \em et al.\egroup
  }{2021}]{hangartner2021monitoring}
Dominik Hangartner, Daniel Kopp, and Michael Siegenthaler.
\newblock Monitoring hiring discrimination through online recruitment
  platforms.
\newblock {\em Nature}, 589(7843):572--576, 2021.

\bibitem[\protect\citeauthoryear{Hardt \bgroup \em et al.\egroup
  }{2016}]{hardt2016equality}
Moritz Hardt, Eric Price, and Nati Srebro.
\newblock Equality of opportunity in supervised learning.
\newblock {\em Advances in neural information processing systems},
  29:3315--3323, 2016.

\bibitem[\protect\citeauthoryear{Iosifidis and
  Ntoutsi}{2019}]{iosifidis2019adafair}
Vasileios Iosifidis and Eirini Ntoutsi.
\newblock Adafair: Cumulative fairness adaptive boosting.
\newblock In {\em Proceedings of the 28th ACM International Conference on
  Information and Knowledge Management}, pages 781--790, 2019.

\bibitem[\protect\citeauthoryear{Iosifidis \bgroup \em et al.\egroup
  }{2019}]{iosifidis2019fae}
Vasileios Iosifidis, Besnik Fetahu, and Eirini Ntoutsi.
\newblock Fae: A fairness-aware ensemble framework.
\newblock In {\em 2019 IEEE International Conference on Big Data (Big Data)},
  pages 1375--1380. IEEE, 2019.

\bibitem[\protect\citeauthoryear{Jacobs and
  Wallach}{2021}]{jacobs2021measurement}
Abigail~Z Jacobs and Hanna Wallach.
\newblock Measurement and fairness.
\newblock In {\em Proceedings of the 2021 ACM conference on fairness,
  accountability, and transparency}, pages 375--385, 2021.

\bibitem[\protect\citeauthoryear{Kotsiantis}{2011}]{Kotsiantis2011}
Sotiris~B. Kotsiantis.
\newblock {Feature selection for machine learning classification problems: A
  recent overview}.
\newblock {\em Artificial Intelligence Review}, 42(1):157--176, 2011.

\bibitem[\protect\citeauthoryear{Kusner \bgroup \em et al.\egroup
  }{2017}]{kusner2017counterfactual}
Matt~J Kusner, Joshua Loftus, Chris Russell, and Ricardo Silva.
\newblock Counterfactual fairness.
\newblock In {\em Advances in Neural Information Processing Systems}, pages
  4066--4076, 2017.

\bibitem[\protect\citeauthoryear{{Le Quy} \bgroup \em et al.\egroup
  }{2022}]{LeQuy2022}
Tai {Le Quy}, Arjun Roy, Vasileios Iosifidis, Wenbin Zhang, and Eirini Ntoutsi.
\newblock {A survey on datasets for fairness-aware machine learning}.
\newblock {\em Wiley Interdisciplinary Reviews: Data Mining and Knowledge
  Discovery}, 12(3):1--59, 2022.

\bibitem[\protect\citeauthoryear{Li \bgroup \em et al.\egroup }{2022}]{Li2022}
Zhu Li, Adrián Pérez-Suay, Gustau Camps-Valls, and Dino Sejdinovic.
\newblock Kernel dependence regularizers and gaussian processes with
  applications to algorithmic fairness.
\newblock {\em Pattern Recognition}, 132:108922, 2022.

\bibitem[\protect\citeauthoryear{Mehrabi \bgroup \em et al.\egroup
  }{2019}]{Mehrabi2019}
Ninareh. Mehrabi, Fred. Morstatter, Nripsuta. Saxena, Kristina. Lerman, and
  Aram. Galstyan.
\newblock A survey on bias and fairness in machine learning.
\newblock {\em arXiv preprint arXiv:1908.09635}, 2019.

\bibitem[\protect\citeauthoryear{Pearl}{2009}]{pearl2009causality}
Judea Pearl.
\newblock {\em Causality}.
\newblock Cambridge university press, 2009.

\bibitem[\protect\citeauthoryear{Pelegrina and Duarte}{2022}]{Pelegrina2022b}
Guilherme~Dean Pelegrina and Leonardo~Tomazeli Duarte.
\newblock {A novel approach for Fair Principal Component Analysis based on
  eigendecomposition}.
\newblock {\em ArXiv ID: 2208.11362}, 2022.

\bibitem[\protect\citeauthoryear{Pelegrina \bgroup \em et al.\egroup
  }{2022}]{Pelegrina2022}
Guilherme~Dean Pelegrina, Renan Del~Buono Brotto, Leonardo~Tomazeli Duarte,
  Romis Attux, and João Marcos~Travassos Romano.
\newblock {Analysis of trade-offs in fair principal component analysis based on
  multi-objective optimization}.
\newblock In {\em 2022 International Joint Conference on Neural Networks
  (IJCNN)}, pages 1--8, Padua, Italy, 2022. IEEE.

\bibitem[\protect\citeauthoryear{Pérez-Suay \bgroup \em et al.\egroup
  }{2017}]{Perez-Suay2017}
Adrián Pérez-Suay, Valero Laparra, Gonzalo Mateo-García, Jordi Muñoz-Marí,
  Luis Gómez-Chova, and Gustau Camps-Valls.
\newblock Fair kernel learning.
\newblock In {\em Machine Learning and Knowledge Discovery in Databases. ECML
  PKDD 2017. Lecture Notes in Computer Science}, volume 10534, pages 339--355.
  Springer, Cham, 2017.

\bibitem[\protect\citeauthoryear{Raji and Buolamwini}{2019}]{Raji2019}
Inioluwa~Deborah Raji and Joy Buolamwini.
\newblock {Actionable auditing: Investigating the impact of publicly naming
  biased performance results of commercial AI products}.
\newblock In {\em Proceedings of the 2019 AAAI/ACM Conference on AI, Ethics,
  and Society}, pages 429--435, 2019.

\bibitem[\protect\citeauthoryear{Roh \bgroup \em et al.\egroup
  }{}]{roh2020fairbatch}
Yuji Roh, Kangwook Lee, Steven~Euijong Whang, and Changho Suh.
\newblock Fairbatch: Batch selection for model fairness.
\newblock In {\em 9th International Conference on Learning Representations,
  {ICLR} 2021}.

\bibitem[\protect\citeauthoryear{Sarker}{2021}]{Sarker2021}
Iqbal~H. Sarker.
\newblock {Machine learning: Algorithms, real-world applications and research
  directions}.
\newblock {\em SN Computer Science}, 2(3):160, 2021.

\bibitem[\protect\citeauthoryear{Sch{\"{o}}lkopf \bgroup \em et al.\egroup
  }{2002}]{Scholkopf2002}
Bernhard Sch{\"{o}}lkopf, Alexander~J. Smola, and Francis Bach.
\newblock {\em {Learning with kernels: Support vector machines, regularization,
  optimization, and beyond}}.
\newblock MIT press, 2002.

\bibitem[\protect\citeauthoryear{Song \bgroup \em et al.\egroup
  }{2007}]{Song2007}
Le~Song, Alex Smola, Arthur Gretton, Karsten~M. Borgwardt, and Justin Bedo.
\newblock Supervised feature felection via dependence estimation.
\newblock In {\em Proceedings of the 24th international conference on Machine
  learning}, pages 823--830, 2007.

\bibitem[\protect\citeauthoryear{Song \bgroup \em et al.\egroup
  }{2012}]{Song2012}
Le~Song, Alex Smola, Arthur Gretton, Justin Bedo, and Karsten Borgwardt.
\newblock Feature selection via dependence maximization.
\newblock {\em Journal of Machine Learning Research}, 13:1393--1434, 2012.

\bibitem[\protect\citeauthoryear{Tsamados \bgroup \em et al.\egroup
  }{2022}]{tsamados2022ethics}
Andreas Tsamados, Nikita Aggarwal, Josh Cowls, Jessica Morley, Huw Roberts,
  Mariarosaria Taddeo, and Luciano Floridi.
\newblock The ethics of algorithms: key problems and solutions.
\newblock {\em AI \& SOCIETY}, 37(1):215--230, 2022.

\bibitem[\protect\citeauthoryear{Vergara and Est{\'{e}}vez}{2014}]{Vergara2014}
Jorge~R. Vergara and Pablo~A. Est{\'{e}}vez.
\newblock {A review of feature selection methods based on mutual information}.
\newblock {\em Neural Computing and Applications}, 24:175--186, 2014.

\bibitem[\protect\citeauthoryear{Wang \bgroup \em et al.\egroup
  }{2020}]{Wang2020}
Hao Wang, Yijie Ding, Jijun Tang, and Fei Guo.
\newblock Identification of membrane protein types via multivariate information
  fusion with hilbert–schmidt independence criterion.
\newblock {\em Neurocomputing}, 383:257--269, 2020.

\bibitem[\protect\citeauthoryear{Wang \bgroup \em et al.\egroup
  }{2021}]{Wang2021}
Tinghua Wang, Xiaolu Dai, and Yuze Liu.
\newblock Learning with hilbert–schmidt independence criterion: A review and
  new perspectives.
\newblock {\em Knowledge-Based Systems}, 234:107567, 2021.

\bibitem[\protect\citeauthoryear{Wightman}{1998}]{Wightman1998}
Linda~F. Wightman.
\newblock {LSAC national longitudinal bar passage study}.
\newblock Technical report, 1998.

\bibitem[\protect\citeauthoryear{Yeh and hui Lien}{2009}]{Yeh2009}
I-Cheng Yeh and Che hui Lien.
\newblock The comparisons of data mining techniques for the predictive accuracy
  of probability of default of credit card clients.
\newblock {\em Expert Systems with Applications}, 36:2473--2480, 2009.

\bibitem[\protect\citeauthoryear{Zemel \bgroup \em et al.\egroup
  }{2013}]{zemel2013learning}
Rich Zemel, Yu~Wu, Kevin Swersky, Toni Pitassi, and Cynthia Dwork.
\newblock Learning fair representations.
\newblock In {\em International Conference on Machine Learning}, pages
  325--333, 2013.

\bibitem[\protect\citeauthoryear{Zhang \bgroup \em et al.\egroup
  }{2018}]{zhang2018mitigating}
Brian~Hu Zhang, Blake Lemoine, and Margaret Mitchell.
\newblock Mitigating unwanted biases with adversarial learning.
\newblock In {\em Proceedings of the 2018 AAAI/ACM Conference on AI, Ethics,
  and Society}, pages 335--340, 2018.

\end{thebibliography}

\end{document}